\documentclass[sigconf]{acmart}

\AtBeginDocument{%
  \providecommand\BibTeX{{%
    \normalfont B\kern-0.5em{\scshape i\kern-0.25em b}\kern-0.8em\TeX}}}

\usepackage{todonotes}
\usepackage{times}
\usepackage{latexsym}
\usepackage{amsmath}
\usepackage{url}
\usepackage{multirow}
\usepackage{graphicx}
\usepackage{graphics}
\usepackage{enumitem}
\usepackage{subcaption}

\usepackage{amssymb}
\usepackage[normalem]{ulem} 
\usepackage{makecell}
\usepackage{array}
\usepackage{float}
\DeclareMathOperator*{\argmax}{arg\,max}

\usepackage{listings}
\usepackage[lined,ruled,commentsnumbered]{algorithm2e}

\usepackage{siunitx}
\copyrightyear{2021}
\acmYear{2021}
\setcopyright{acmlicensed}\acmConference[KDD '21]{Proceedings of the 27th ACM SIGKDD Conference on Knowledge Discovery and Data Mining}{August 14--18, 2021}{Virtual Event, Singapore}
\acmBooktitle{Proceedings of the 27th ACM SIGKDD Conference on Knowledge Discovery and Data Mining (KDD '21), August 14--18, 2021, Virtual Event, Singapore}
\acmPrice{15.00}
\acmDOI{10.1145/3447548.3467202}
\acmISBN{978-1-4503-8332-5/21/08}

\begin{document}

\title{Diversity driven Query Rewriting in Search Advertising}

\author{Akash Kumar Mohankumar}
\authornote{Both authors contributed equally to this work}
\authornote{Work done during internship at Microsoft}
\email{makashkumar99@gmail.com}
\orcid{1234-5678-9012}
\affiliation{%
  \institution{Indian Institute of Technology Madras}
  \city{Chennai}
  \country{India}
}

\author{Nikit Begwani}
\authornotemark[1]
\email{nibegwan@microsoft.com}
\affiliation{%
  \institution{Microsoft}
  \city{Bangalore}
  \country{India}
  }
  
\author{Amit Singh}
\email{siamit@microsoft.com}
\affiliation{%
  \institution{Microsoft}
  \city{Bangalore}
  \country{India}
  }








\begin{abstract}
Retrieving keywords (bidwords) with the same intent as query, referred to as close variant keywords, is of prime importance for effective targeted search advertising. For head and torso search queries, sponsored search engines use a huge repository of same intent queries and keywords, mined ahead of time. Online, this repository is used to rewrite the query and then lookup the rewrite in a repository of bid keywords contributing to significant revenue. Recently generative retrieval models have been shown to be effective at the task of generating such query rewrites. We observe two main limitations of such generative models. First, rewrites generated by these models exhibit low lexical diversity, and hence the rewrites fail to retrieve relevant keywords that have diverse linguistic variations. Second, there is a misalignment between the training objective - the likelihood of training data, v/s what we desire - improved quality and coverage of rewrites. In this work, we introduce CLOVER, a framework to generate both high-quality and diverse rewrites by optimizing for human assessment of rewrite quality using our diversity-driven reinforcement learning algorithm. We use an evaluation model, trained to predict human judgments, as the reward function to finetune the generation policy.  We empirically show the effectiveness of our proposed approach through offline experiments on search queries across geographies spanning three major languages. We also perform online A/B experiments on Bing, a large commercial search engine, which shows (i) better user engagement with an average increase in clicks by 12.83\% accompanied with an average defect reduction by 13.97\%, and (ii) improved revenue by 21.29\%. 
   
   
\end{abstract}

\keywords{sponsored search, query rewriting, natural language generation}


\maketitle
\section{Introduction}
Sponsored search is an important component of modern search engines and accounts for a significant portion of their revenue. In sponsored search, advertisers bid on keywords relevant to their business to place their ads along with the organic search results. The search engine must match a user's search query to relevant keywords. Traditionally, search engines only used exact match where a user's search query is matched to a bid keyword only if they are exactly the same. The exact match type poses a massive challenge for advertisers to enumerate all possible variations of a keyword and separately bid on each of them. Further, exact match type leads to potential revenue loss to advertisers and the search engine since ads aren't displayed even if a user's query is only a minor modification of a bid keyword. To overcome these problems, search engines now provide extensions to exact match such as close-variant match types \footnote{https://support.google.com/google-ads/answer/9342105?hl=en} where search queries are matched to bid keywords with the same search intent even though they are expressed with different words. This problem of matching a search query to its close-variant keywords is challenging for many reasons. First, the query-keyword match should be of high quality, \textit{i.e.} the matched keywords should strictly have the same search intent as the original search query. Second, the matching algorithm should retrieve all possible high-quality bid keywords for a given query, resulting in the maximum coverage from the bid keyword library. Third, the matching algorithms must be multilingual since modern search engines support search in various languages. \\

Matching search queries with bid keywords online in real-time places latency and compute constraints on the matching algorithm, often resulting in lower quality and coverage. Therefore in addition to online rewrites, search engines mine bid keywords offline for frequent queries and store them ahead of time in a production table, which is a multivalue map consisting of entries of the form \{\textit{${q: k_1, k_2 \dots k_m}$}\} where $q, k$ represent queries and keywords respectively. Such offline close-variant matching can be performed using various approaches in the literature. One standard method is to use Information Retrieval (IR) algorithms to retrieve relevant keywords for a given search query and then apply quality filters \cite{quality_filter} to remove the low-quality keywords. A common problem observed in IR methods, dubbed as the "semantic gap" problem \cite{baidu_generative_trie, ProphetNetAds}, is IR methods often fail to retrieve many semantically relevant keywords for a given query. Few works \cite{query_expansion_1, query_expansion_2, query_expansion_3} try to mitigate this problem by rewriting the input search query into several intermediate queries and then combine the retrieved keywords obtained from all the intermediate queries. However, as pointed out by \cite{baidu_generative_trie}, such a multi-stage procedure can result in a gradual accumulation of errors in each sub-module resulting in low precision and recall rates. Recent methods \cite{baidu_generative_trie, ProphetNetAds} focus on leveraging advancements in Natural Language Generation (NLG) by viewing this problem as a constrained generation task. Such generative approaches first involve training language models to generate query rewrites in a supervised setup. Later, during inference with beam search, the generations are constrained to the bid keyword library using a prefix trie. \\

\begin{figure}
    \centering
    \includegraphics[width=0.45\textwidth]{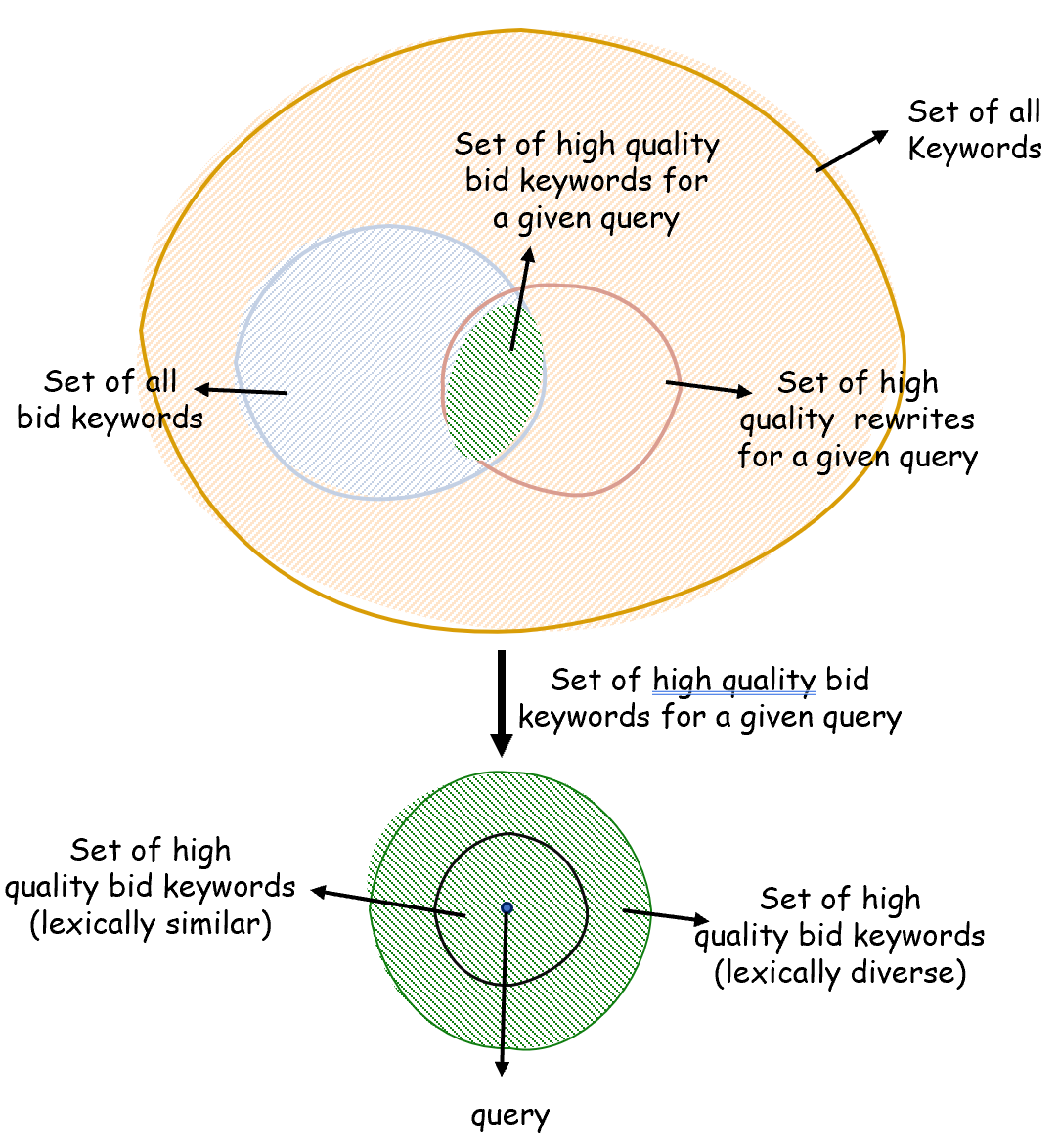}
    \caption{Diagram of different possible keyword sets}
    \label{fig:venndiagram}
\end{figure}

Despite the performance improvement over IR baselines, existing generative approaches have a few important shortcomings. Firstly, rewrites generated with beam search for a given input query tend to be very similar to each other. For example, as shown in the column 2 of table \ref{tab:DiversityExample}, the generated rewrites are only minor syntactical variations of the input query such as modifying word order, adding articles, etc. Such low linguistic diversity in the rewrites leads to limited coverage as we fail to retrieve several other keywords that have diverse lexical variations. Another problem with these approaches is the log-likelihood training objective is only a rough proxy of what we actually care about - improved quality and coverage in the rewrites. For instance, the log-likelihood objective cannot distinguish between important (e.g., adding irrelevant words in the rewrite) and unimportant (e.g., selecting the precise word from a list of synonyms) errors and penalize the model accordingly. Further, the maximum likelihood approach encourages the model to increase the probability of generating the training data even if it is of poor quality. Also, the training dataset, consisting of queries and its corresponding rewrites, itself is often limited. For example, the training data usually only contains a few rewrites for each query, but our objective is to generate all possible relevant keywords for a search query during inference to achieve high coverage. \\

\begin{table*}[h]
\begin{center}
 \begin{subtable}[h]{\textwidth}
    \small\addtolength{\tabcolsep}{-4pt}
    \scalebox{1}{
	\begin{tabular}{c c c c c }	
		\hline
		\textbf{Query} & \textbf{Beam Search} & \textbf{BS with Diversity}  & \textbf{BS with Diversity}  & \textbf{Our Diversity RL}\\ 
		& \textbf{(BS)} & (High Quality) & (Quality Drift) & (Drift Controlled)\\ \hline
		&accommodations London & hotel accommodation london & \textbf{\textcolor{red}{travel}} London & \textbf{\textcolor{black}{shared}} accomodation london\\
		\textit{accommodations in london}&accommodations in London  & houses in london  & \textbf{\textcolor{red}{flights}} from London & \textbf{\textcolor{black}{best hotels}} london\\ 
		
		&london accommodation & hostels london  	& London \textbf{\textcolor{red}{tourism}} & \textbf{\textcolor{black}{lodging}} in london \textbf{\textcolor{black}{uk}}\\ \hline
		
		&consistent & headache in a child &  consistent \textbf{\textcolor{red}{weight gain}} & \textbf{\textcolor{black}{what is the best}} headache \textbf{\textcolor{black}{medicine}} \\
		\textit{consistent headache} & headache constant & headache fasting & consistent \textbf{\textcolor{red}{heartburn}} & headache \textbf{\textcolor{black}{concussion}} \\ 
		&consistent muscle pain & chronic headache and dizziness 	& consistent \textbf{\textcolor{red}{options income}} & headache \textbf{\textcolor{black}{in one side}} \\ \hline
		
		&the best compact cars & best compact cars used & the top car \textbf{\textcolor{red}{insurance companies}} & \textbf{\textcolor{black}{top}} compact cars \textbf{\textcolor{black}{2020}} \\
		\textit{best compact car}&compact car  &what is the best compact car &compact \textbf{\textcolor{red}{compacts powder}} & \textbf{\textcolor{black}{most}} compact car \textbf{\textcolor{black}{seat}}\\ 
		&best compact vehicles & best compact cars in snow & which best compact \textbf{\textcolor{red}{camera}} & \textbf{\textcolor{black}{affordable}} compact cars \\ \hline
		
		&increase wifi & how to improve wifi range & \textbf{\textcolor{red}{exchange}} wifi & \textbf{\textcolor{black}{aumentare portata}} wifi\\
		\textit{increase wifi range}&increase wireless range  & increase my wifi signal& \textbf{\textcolor{red}{decreased value accident}} & \textbf{\textcolor{black}{upgrade}} wifi \textbf{\textcolor{black}{speed}}\\ 
		&increase wifi power & enhance wifi & increased internet speed & wifi range \textbf{\textcolor{black}{booster}}\\ \hline
	\end{tabular}	
	}
    \end{subtable}

	\caption{Samples of generated keywords by different methods on four search queries}
	\label{tab:DiversityExample}	
\end{center}
\end{table*}

As illustrated in figure \ref{fig:venndiagram}, the set of all high-quality bid keywords includes both lexically similar and diverse keywords. In this work, we specifically focus on generating such lexically diverse keywords that are more challenging to be retrieved. We empirically show that adopting a diverse decoding algorithm \cite{DSS} during inference can partially overcome the problem with standard beam search. For example, as shown in column 3 of table \ref{tab:DiversityExample}, we are now able to retrieve keywords with greater linguistic diversity. However, we also observe instances where with increased diversity, the generated rewrites drift away from the original search intent (column 4 of table \ref{tab:DiversityExample}). In this work, we show that it is possible to retrieve keywords with high-quality and high lexical diversity that were not previously retrieved using our Close Variant Generation (CLOVER) framework. We propose a more principled approach to overcome the previously discussed problems by directly optimizing human judgments on rewrite quality with a controllable diversity constraint. We first train an evaluation model to predict human judgments on rewrite quality of generated rewrites. We then use the evaluation model as a reward function to finetune the generation policy in a reinforcement learning setup. However, we observe a crucial problem with the standard RL objective. We show that it is sufficient for the generation policy to place a high probability mass one plausible rewrite and ignore all other valid rewrites to obtain high rewards in the standard RL setup. To overcome this problem, we propose our diversity driven RL algorithm where the optimization objective enforces generating multiple diverse and high-quality rewrites. As shown in column 5 of table \ref{tab:DiversityExample}, our proposed approach generates rewrites with high-quality and high diversity. We perform offline experiments on queries from a large commercial search engine spanning three languages (English, German and French) and demonstrate the effectiveness of our proposed approach. We further conducted online experiments on a commercial search engine\footnote{Microsoft Bing}across regions spanning three major languages and obtained an average of 12.83\% improvement in clicks, 13.97\% reduction in defect, and 21.29\% improvement in revenue from close variants. To sum up, the key contributions of our work are listed below:
\begin{itemize}
    \item We identify the key problems limiting the performance of existing generative retrieval approaches in sponsored search. We empirically show that some of these problems can be partially alleviated using a diverse decoding algorithm. 

    \item We propose CLOVER, a framework for optimizing human judgments on rewrite quality while also being able to control the desired diversity. 
    
    \item Through offline experiments on search queries, we show that our proposed approach generates diverse and high-quality keywords that were not retrieved by previous methods. 
    
    \item Finally, we conduct online A/B experiments on a commercial search engine and show that our proposed methods leads to significantly better user engagement and revenue gains.

\end{itemize}
\section{Related Work}
\label{sec:related_work}
{\flushleft \textbf{Sponsored Search: }}
Early works in sponsored search \cite{Broder2007ASA, Broder2008SearchAU, RibeiroNeto2005ImpedanceCI} mainly focused on matching queries with relevant advertisements using standard information retrieval algorithms where an advertisement is treated as a document consisting of textual information such as the bid keyword, ad text visible to users, etc. Such approaches mainly use bag-of-words features like tf-idf to match queries with ads. Some other works \cite{Bai2018ScalableQN, Grbovic2016ScalableSM} learn embeddings of queries and ads in a shared vector space from search session, ad click, and search link click data using word2vec \cite{word2vec} like algorithms. A few recent works \cite{Li2019LearningFM} exploit advances in neural information retrieval models \cite{Mitra2017NeuralMF} such as Deep Crossing \cite{Shan2016DeepCW} in sponsored search. An important line of work \cite{query_expansion_1, query_expansion_2, query_expansion_3, Grbovic2015ContextAC, Malekian2008OptimizingQR} that is based on the idea of performing query to query transformations, also known as query rewriting. In these methods, the input search query is rewritten to multiple intermediate queries, and then the bid keywords are retrieved from all the intermediate queries either with exact match or more advanced matching algorithms. Another related area of research \cite{Azimi2015AdsKR, Zhou2019DomainConstrainedAK} involves keyword to keyword transformations referred to as keyword rewriting, where the objective is to rewrite less frequent bid keywords without altering the search original intent. Recently, there is a greater focus on direct query to keyword transformations \cite{baidu_generative_trie, ProphetNetAds, Lee2018RareQE}. In these approaches, the input query is directly converted to several bid keywords using seq2seq language models with trie based decoding techniques to constraint the output space during inference. Another recent work \cite{Dahiya2019DeepXMLS} views matching bid keywords to queries as an extreme multi-label classification problem and propose deep extreme multi-label learning algorithms. 
{\flushleft \textbf{Reinforcement Learning in NLP: }} In NLP, there is a rich literature in using reinforcement algorithms to optimize heuristic-based evaluation metrics such as BLEU in machine translation \cite{Ranzato2016, Nguyen2017ReinforcementLF} and ROUGE in automatic summarization \cite{Ranzato2016, Wu2018LearningTE}. Such works usually use policy gradient algorithms such as REINFORCE with baseline \cite{REINFORCE} to optimize the rewards where the baseline can either be learned from data \cite{Ranzato2016} or can be set to the reward obtained from greedy decoding \cite{SelfCriticalST}. There have also been works \cite{Bahdanau2017AnAA} that have investigated Actor-Critic algorithms \cite{Konda1999ActorCriticA} in NLP. Recently, there has been an increasing focus on using rewards from trained evaluation models in various tasks such as Machine Translation \cite{Kreutzer2018CanNM}, Abstracting Summarization \cite{Bhm2019BetterRY, summarization_from_human_feedback, FineTuningLM}. Unlike the previous works which focus on generation quality, our work focuses on improving both generation quality and diversity. To this end, we propose a novel RL algorithm where the performance measure of a policy is defined as the expected rewards from multiple diverse samples decoded from the policy and not just one sample as defined in the previous works. 

\begin{figure*}
    \centering
    \includegraphics[width=0.8\textwidth]{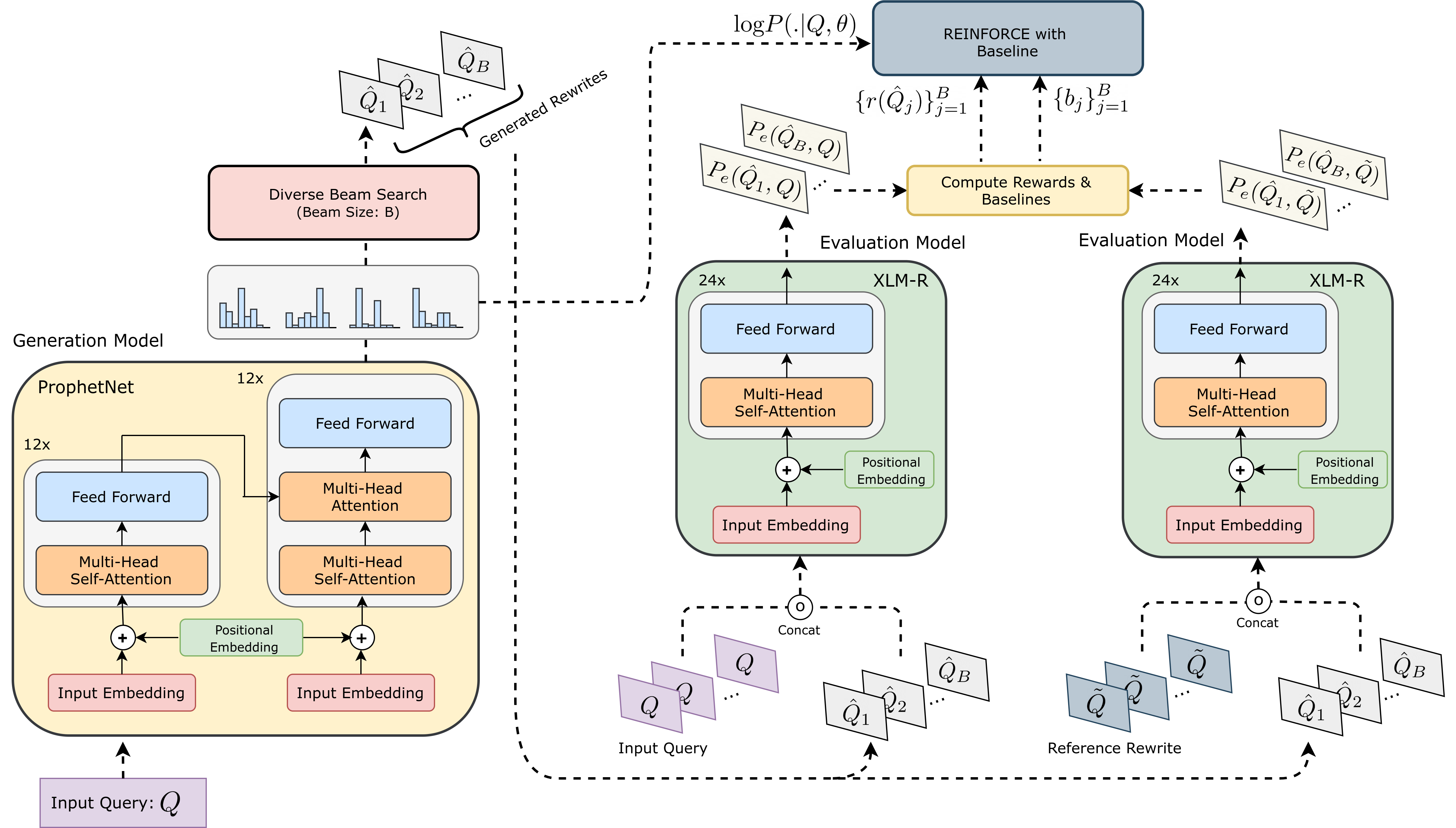}
    \caption{CLOVER: Consisting of a Generator and Evaluation model with the diversity driven RL algorithm}
    \label{fig:model}
\end{figure*}

\section{Methodology}
In this section, we discuss the various components of our Close Variant Generation (CLOVER) framework, as outlined in figure \ref{fig:model}. Given a user's search query $Q$, our objective is to retrieve all keywords $K$ with the same search intent as $Q$ from the set of bid keywords $\mathcal{K}$. We first discuss the details of our language generation model in section \ref{generative_models}. Next, in section \ref{evaluation_model}, we discuss our evaluation model, which is trained to evaluate the quality of rewrites. Later in \ref{standard_rl}, we discuss the standard RL approach used to finetune the generation model with rewards. We then discuss the limitations of the standard RL formulation and propose our Diversity Driven RL approach in \ref{diversity_driven_rl}. Lastly, in \ref{decoding_algorithms}, we discuss the decoding algorithms. 

\subsection{Generation Model}
\label{generative_models}
For generating query rewrites, we adopt the ProphetNet model \cite{prophetnet} which has been shown to obtain state-of-the-art performance on several seq2seq language generation tasks. The model follows the standard transformer-large encoder-decoder architecture \cite{attention_is_all_need} with 12 layers of encoder and 12 layers of decoder. We use a multi-lingual version of the ProphetNet model that is pretrained on the multilingual (100 languages) Wikipedia corpus using the token span masking objective \cite{MASS}. We then finetune the generation model for query rewriting using the the next n-gram prediction objective \cite{prophetnet} in a supervised manner \textit{i.e.} the model is trained to predict the next n tokens of the rewrite given the previous tokens and the input query. More specifically, given a dataset of query rewrite pairs $\mathcal{D} = \{Q^{(i)}, \Tilde{Q}^{(i)}\}_{i=1}^{m}$, where $Q^{(i)} = \{ w_1^{(i)}, \dots, w_m^{(i)}\}$, $\Tilde{Q}^{(i)} = \{  \Tilde{w}_1^{(i)}, \dots, \Tilde{w}_{\Tilde{m}}^{(i)}\}$, our training objective is as follows: 
$$
    \mathcal{L}(\theta, \mathcal{D}) = - E_{Q, \Tilde{Q} \sim \mathcal{D}} \left[ \sum_{j=0}^{n-1} \alpha_j \left( \sum_{t=1}^T \mbox{log} p(\Tilde{w}_{t+j}|\Tilde{w}_{<t}, Q, \theta) \right) \right]
$$

where $\theta$ represents the model parameters, $\{\alpha_j\}_{j=0}^{n-1}$ are the weights given to each future token prediction. During training, all the future n tokens are simultaneously predicted with the help of an n-stream self-attention mechanism. During inference, only the mainstream self-attention is used to predict the tokens one at a time.
\subsection{Evaluation Model}
\label{evaluation_model}
The goal of the evaluation model is to judge the quality of a generated rewrite $\hat{Q}$ corresponding to an input query $Q$. Several recent studies \cite{nlg_eval_survey, nlg_eval_study_1,nlg_eval_study_3,Sai2020ImprovingDE} have shown that heuristic-based evaluation metrics that rely on n-gram overlap \cite{bleu, rouge, meteor} or word embedding similarity \cite{embedding_average, greedymatch, vectorextrema} with a reference have very poor correlations with human judgements on several NLG tasks. We observed that the performance of such heuristic-based metrics in query rewriting is indeed abysmal. To illustrate this, consider an input query "sennheiser headphones with attached microphone" with a reference rewrite "sennheiser headphones attached with microphone". The rewrite "sony headphones attached with microphone" is exactly the same as the reference rewrite except for one word but does not have the same search intent as the input. On the other hand, the rewrite "sennheiser headsets" has only a single word in common with the reference but is a valid rewrite. Such nuances in search intent cannot be handled by metrics that rely on simple hand-crafted rules. Therefore, we develop an end-to-end evaluation metric specifically trained to evaluate the quality of generated rewrites. We use the XLM-RoBERTa (XLM-R) \cite{xlmr} model with 24 transformer encoder layers that is pre-trained on 100 languages. We fine-tune the XLM-R model on a dataset consisting of triplets of the form $(Q, \hat{Q}, y)$ where $y$ represents the human judgments provided by annotators about quality of the rewrite $\hat{Q}$. The scores provided by the trained evaluation model, denoted by $P_e(\hat{Q}, Q)$ is used to evaluate the quality of the generated rewrites and is also used as rewards to the RL algorithms as we described in the next subsection. 
\subsection{Reinforcement Learning Approach}
\label{standard_rl}

Query rewriting can be formulated as a reinforcement learning problem, similar to other sequence generation problems \cite{Ranzato2016, SelfCriticalST, FineTuningLM, summarization_from_human_feedback}: At the start of an episode, the environment chooses a search query $Q$ (initial state). The agent (generator) chooses an action $\hat{w}_{t}$ (next word) at time $t$ according to the policy $P(. | \hat{w}_{<t}, Q, \theta)$. After taking an action, the agent updates its internal state and continues rolling out the policy until the end token or maximum sequence length is reached. The environment then provides a reward for the generated rewrite $\hat{Q}$ (sequence of actions), which the agent aims to optimize. Formally, the goal of the agent is to learn the policy parameter $\theta$ that maximizes: 
\begin{equation}
    \label{eqn:rl_cost}
    J(\theta) = E_{Q} E_{\hat{Q} \sim P(.| Q, \theta)} r(\hat{Q})
\end{equation}

where $\hat{Q} = \{ \hat{w}_1, \dots  \hat{w}_m\}$, $\hat{w}_{t}$ is the sampled token from the policy $P(. | \hat{w}_{<t}, Q, \theta)$ at time $t$,  $r(\hat{Q})$ is the reward for $\hat{Q}$ which can depend on the search query $Q$ and optionally a reference rewrite $\Tilde{Q}$ (these dependencies are omitted from the notation for brevity). We now give a brief overview of the standard policy gradients algorithms used to optimize the objective in equation \ref{eqn:rl_cost}. 

{\flushleft \textbf{REINFORCE:}} The REINFORCE algorithm \cite{REINFORCE} uses an unbiased gradient estimator to approximate the exact gradients with Monte-Carlo samples. Given a mini batch of search queries $\{Q^{(i)}\}_{i=1}^{n}$ and the corresponding rewrites $\{\hat{Q}^{(i)}\}_{i=1}^{n}$ sampled from the policy $P(. | Q, \theta)$, the gradient is approximated as follows:
\begin{equation}
    \label{eqn:reinforce}
    \nabla_{\theta} J(\theta) \approx \sum_{i=1}^n r(\hat{Q}^{(i)}) \nabla_{\theta} \mbox{log} P(\hat{Q}^{(i)} | Q^{(i)}, \theta)
\end{equation}

{\flushleft \textbf{REINFORCE with baseline:}} Although, the REINFORCE gradient estimator is unbiased, it suffers from very high variance. A common technique used in practice to reduce the variance is to add a baseline function $b$ to the gradient estimator: 

\begin{equation}
    \nabla_{\theta} J(\theta) \approx \sum_{i=1}^n  \left(r(\hat{Q}^{(i)}) - b^{(i)} \right) \nabla_{\theta} \mbox{log} P(\hat{Q}^{(i)} | Q^{(i)}, \theta)
\end{equation}

It can be shown that the new gradient estimator is unbiased if the baseline $b^{(i)}$ does not depend on the sampled actions $\hat{Q}^{(i)}$ \cite{REINFORCE}. 

{\flushleft \textbf{Reward function:}} We compute the scores provided by our evaluation model for a generated rewrite $\hat{Q}$ by comparing it with the the input query $Q$ and with the reference rewrite $\Tilde{Q}$. We use the weighted combinations of these two scores as our reward function. Mathematically, the reward function is defined as follows: 
\begin{equation}
    r(\hat{Q}) = \alpha P_e(\hat{Q}, Q) + (1-\alpha) P_e(\hat{Q}, \Tilde{Q})
\end{equation}
If high-quality reference rewrites are not available, we can set $\alpha$ to 1 and neglect the  $P_e(\hat{Q}, \Tilde{Q})$ term. 

\subsection{Diversity driven Reinforcement Learning}
\label{diversity_driven_rl}
A crucial problem with the standard reinforcement learning setup is that in order to optimize equation \ref{eqn:rl_cost}, it is sufficient for the policy to place a high probability mass on one plausible rewrite and ignore all other valid rewrites for a given input query. Concretely, there always exists an optimal policy $P^*(\hat{Q} | Q)$ which is completely deterministic given the input query \textit{i.e.} it generates only one rewrite per query:  
\[    P^* (\hat{Q} | Q) = 
    \begin{cases}
        1,& \text{if } \hat{Q} = \argmax_{\hat{Q}} {r(\hat{Q})}\\
        0              & \text{otherwise}
    \end{cases}
\]

In fact, we shall show in section \ref{sec:offline_results} that as we finetune the generator model using the REINFORCE with baseline algorithm, the entropy of the policy keeps decreasing and approaches a very low value. The key problem here lies with the standard RL objective since it only requires expected rewards of one sample (rewrite) from the policy to be high. Unlike other generation tasks such as machine translation, this is a significant problem in query rewriting since our primary objective is to generate as many distinct high-quality rewrites as possible for a given search query. 

To overcome these issues in the standard RL objective, we propose our diversity driven reinforcement learning problem formulation. Our formulation uses a diverse decoding subroutine $\mathcal{M}$ that takes as input the generator model $P(.|Q, \theta)$, its parameter $\theta$, an input query $Q$, a diversity parameter $\gamma$ and returns $B$ diverse samples $\{\hat{Q}_1, \dots, \hat{Q}_B\}$ without replacement from the policy $P$, where $\hat{Q}_j = \{\hat{w}_{j1} \dots, \hat{w}_{jm}\}$. We shall discuss more details about such decoding algorithms in the next subsection. In diversity driven RL, our goal is to maximize the expected total rewards obtained from $B$ diverse rewrites sampled from the policy using the decoding subroutine $\mathcal{M}$. Formally, our objective is defined as follows:
\begin{align}
        J(\theta, \mathcal{M}) &= E_{Q} J(\theta, Q, \mathcal{M}) \label{eqn:diverserl_objective} \\
        J(\theta, Q, \mathcal{M}) &= \sum_{j=1}^{B}
        E_{\hat{Q}_{j} \sim \mathcal{M}(P, \theta, Q, \gamma)} r(\hat{Q}_{j})
\end{align}
The main difference here is that optimizing the performance measure in equation \ref{eqn:diverserl_objective} results in policies that when decoded with $\mathcal{M}$ produce both diverse and high quality rewrites. The amount of desired diversity in the rewrites can be controlled by the diversity parameter $\gamma$. To optimize the new objective, we follow the REINFORCE with baseline algorithm and approximate the gradient as:  
\begin{align}
        \nabla_{\theta} J(\theta, \mathcal{M})  &\approx \sum_{i=1}^{n} \sum_{j=1}^{B} \left( (r(\hat{Q}^{(i)}_j) - b^{(i)}_j) \nabla_{\theta} \text{log} P(\hat{Q}^{(i)}_j | Q^{(i)}, \theta) \right) \\
        b^{(i)}_j &= \frac{1}{B-1} \sum_{k=1, k \neq j}^{B} r(\hat{Q}^{(i)}_k)
\end{align}
where $\{Q^{(i)}\}_{i=1}^{n}$ is a mini-batch of search queries, $\{ \{\hat{Q}^{(i)}_{1} \dots \hat{Q}^{(i)}_{B} \} \}_{i=1}^{n}$ is the corresponding rewrites decoded using  $\mathcal{M}(P, \theta, Q^{(i)}, \gamma)$. As baseline for a rewrite, we use the average rewards obtained by all rewrites for the same search query other than itself. The overall training procedure is shown in algorithm \ref{algo_rl}.

\subsection{Decoding Algorithm}
\label{decoding_algorithms}
We now discuss in detail about the decoding algorithm $\mathcal{M}$, described in the previous sub-section, for sampling diverse rewrites from the generator. We experimented with various possible choices such as stochastic beam search \cite{stochastic-beam-search} and other deterministic approximations: diverse beam search \cite{dbs} and diverse sibling search \cite{DSS}. In practice, we found the diverse sibling search algorithm to work the best. Diverse sibling search is a variant of the standard beam search algorithm that penalizes decoding similar hypothesis within the beam. The standard beam search algorithm consists of selecting the top B hypothesis with the maximum sum of log probability $S(\hat{w}_{t}, \hat{w}_{<t} | Q)$ at each time step $t$ where $S(\hat{w}_{t}, \hat{w}_{<t} | Q) = S(\hat{w}_{<t} | Q) + \log  {p}(\hat{w}_t| \hat{w}_{<t}, Q, T_{\mathcal{K}})$. Diverse sibling search adds a penalty term that discourages multiple expansions $\hat{w}_t$ with the same prefix $\hat{w}_{<t}$. The modified beam search objective is $\Tilde{S}(\hat{w}_{t}, \hat{w}_{<t} | Q) = S(\hat{w}_{t}, \hat{w}_{<t} | Q) - \gamma k$ where $k$ is the rank of the current hypothesis $[\hat{w}_{t}, \hat{w}_{<t}]$ among its siblings that have the same prefix $\hat{w}_{<t}$. $\gamma$ is the diversity factor that controls the desire degree of diversity 

During inference, we constrain our decoding space with a prefix trie $T_{\mathcal{K}}$ of sub-word tokens consisting of all the keywords in the bid keyword library $\mathcal{K}$. When decoding a sub-token $\hat{w}_t$ at time step $t$, we prune all words $w$ that are not a suffix of the already generated sequence $\hat{w}_{<t}$ in the trie. More formally, the (unnormalized) constrained probability distribution is given by:
\begin{equation*}
    \Tilde{p}(\hat{w}_t = w| \hat{w}_{<t}, Q) = 
    \begin{cases}
    p(\hat{w}_t = w| \hat{w}_{<t}, Q),& \text{if } w\in \text{suffix}_{T_{\mathcal{K}}}(\hat{w}_{<t})\\
    0,              & \text{otherwise}
\end{cases}
\end{equation*}
We use the decoding algorithm $\mathcal{M}$ with this constrained distribution during inference to ensure that all the rewrites are valid bid keywords. 
\SetKwInput{KwInput}{Input}
\SetKwInput{KwOutput}{Output}
\begin{algorithm}
    \SetKwData{Left}{left}\SetKwData{This}{this}\SetKwData{Up}{up}
    \SetKwFunction{Union}{Union}\SetKwFunction{FindCompress}{FindCompress}
    \SetKwInOut{Input}{Input}\SetKwInOut{Output}{Output}
    
    \Input{Train dataset $\mathcal{D}$, batch size $bs$, diverse decoding sub-routine $\mathcal{M}$, beam size $B$, diversity strength $\gamma$, reward weight $\alpha$}
    
    \Output{Learned Generator parameters $\theta$}
    \BlankLine
    \For{number of training iterations}
    {
        Sample $\{ ({Q}^{(1)}, {\Tilde{Q}}^{(1)}) \dots ({Q}^{(bs)}, {\Tilde{Q}}^{(bs)})\} \sim \mathcal{D}$ \;
        \For{$i\leftarrow 1$ \KwTo $bs$}
        {   
            $\{\hat{Q}^{(i)}_1, \dots, \hat{Q}^{(i)}_B\} \leftarrow \mathcal{M}(P, \theta, Q^{(i)}, B, \gamma)$ \;
            \For{$j\leftarrow 1$ \KwTo $B$}
            {
                $r(\hat{Q}^{(i)}_j) \leftarrow \alpha P_e(\hat{Q}^{(i)}_j, Q^{(i)}) + (1-\alpha) P_e(\hat{Q}^{(i)}_j, \Tilde{Q}^{(i)})$
            }
            \For{$j\leftarrow 1$ \KwTo $B$}
            {
                $b^{(i)}_j \leftarrow \frac{1}{B-1} \sum_{k=1, k \neq j}^{B} r(\hat{Q}^{(i)}_k)$
            }
        }
        Update the generator by ascending its gradient $(\nabla_{\theta}J)$:
        $\sum_{i=1}^{bs} \sum_{j=1}^{B} \left( (r(\hat{Q}^{(i)}_j) - b^{(i)}_j) \nabla_{\theta} \text{log} P(\hat{Q}^{(i)}_j | Q^{(i)}, \theta) \right)$\;
    }
    \caption{Training Procedure for Learning Generator Parameters with Diversity driven Reinforcement Learning}\label{algo_rl}
\end{algorithm}\DecMargin{1em}

\section{Offline Experiments}
\label{offline_experiments}

\subsection{Dataset Description}
In this section, we describe the various datasets used for training and evaluating (i) the Generation model and (ii) the Evaluation model. All our datasets are obtained from search queries across geographies that span three main languages: English, French and German.

{\flushleft \textbf{Generation Models:}}
For training our generation models, we used two training sets dubbed Ads-large and Ads-small. Ads-large is a massive dataset of 330 million pairs of queries and corresponding rewrites mined from the search logs of a commercial search engine. We finetuned our generator model in a supervised setup using the Ads-large dataset. Training in the RL setup is considerably slower than in the supervised setup with teacher forcing because the output tokens are decoded one word at a time in an auto-regressive manner. Therefore, we used a smaller training set for RL finetuning called Ads-small, 
which consists of 24 million query rewrite pairs. For evaluating the generator  models, we carefully curated a dataset of 80k search queries. The average length of the queries in these datasets is around 3.42 words. 

{\flushleft \textbf{Evaluation Model:}}
To finetune our evaluation model, we collected a dataset of human judgments on rewrite quality. We showed human annotators the original query and a generated rewrite and asked them to rate the rewrite on a 5 point Likert scale with scores from 1 to 5 representing bad, fair, good, perfect, and excellent, respectively. We obtained annotations from multiple annotators for each query rewrite pair to arrive at a final consensus over the rating. We collected such annotations for 75k query rewrite pairs (45k in English and 15k in French and German each) where the rewrites were obtained from various query rewriting models in production.

\subsection{Experimental Details}
\textbf{Generator Supervised Finetuning: } We finetuned the ProphetNet model from the pretrained checkpoint for 5 epochs on the Ads-large dataset. Following \cite{prophetnet}, we use $n=2$ and $\alpha_1 = 0.5, \alpha_2 = 0.5$ in the next n-gram prediction objective. We set the dropout to 0.1, weight decay to 0.01 and label smoothing to 0.1. We used a batch size of 4096 tokens with ADAM optimizer ($\beta_1 = 0.9, \beta_2=0.999$) and learning rate of 1e-5. We used mixed-precision training on 8 Nvidia 16Gb V100 GPUs. The overall training time is close to two weeks. 

{\flushleft \textbf{Generator RL Finetuning:}}
We further finetune the ProphetNet model from the supervised checkpoint for 1 epoch on the Ads-small dataset. We used a batch size of 224 tokens and a beam size of 20 during training and used 4 Nvidia 32Gb V100 GPUs. The remaining details are the same as in the supervised setup. 

{\flushleft \textbf{Evaluation Model Finetuning:}}
We finetuned our evaluation model from the publicly released XLM-R pre-trained checkpoint on our dataset of human judgments. We converted the human ratings into binary labels were ratings of 3 and above \textit{i.e.} good and above ratings were converted to a label of 1 (positive), and the rest were given a label of 0 (negative). We used binary cross-entropy as our loss function with the binary human labels as the targets. 

\begin{table*}[h]
\begin{center}
	\begin{tabular}{c c c c c c}		
		\hline
		\textbf{Finetuning} & \textbf{Decoding Algorithm} &\textbf{Unique High Quality} & \textbf{Distinct Unigram} & \textbf{Distinct Bigram} \\
		\textbf{Approach}& &\textbf{ Bid Keyword} & & \\ \hline
Supervised & BS w/o Trie \textbf{(I)} & 2.10 & {0.209} & {0.573} \\ \hline
Supervised & Normal BS with Trie \textbf{(II)}  & 3.80 & {0.590} & {0.868} \\
Supervised & Diverse Sibling Search with Trie \textbf{(III)} & 3.84 & {0.591} & {0.869}  \\
Diverse RL & Diverse Sibling Search with Trie \textbf{(IV)} & 3.13 & {\textbf{0.738}} & {\textbf{0.917}}  \\ \hline
\multicolumn{5}{c}{\textbf{Ensemble Results}} \\ \hline
 & Ensemble (I)  &  2.10 & - & -  \\ 
 & Ensemble (I) + (II)  & 5.13 (+3.03) & - & -  \\ 
 & Ensemble (I) + (II) + (III) & 5.52 (+0.39) & - & -  \\ 
 & Ensemble (I) + (II) +(III) + (IV)  & \textbf{7.68 (+2.16)} & - & -  \\ \hline
	\end{tabular}	
	\caption{Offline results on various approaches along with the ensembles}
	\label{tab:OfflineResults}	
\end{center}
\end{table*}
\begin{table*}
\begin{tabular}{c|c|c|c|c|c}
\hline
\multirow{2}{*}{\begin{tabular}[c]{@{}c@{}}\textbf{Decoding Algorithm}\end{tabular}} & \multirow{2}{*}{\begin{tabular}[c]{@{}c@{}}\textbf{Diversity Strength}\end{tabular}} & \multicolumn{2}{c|}{\textbf{Distinct Unigram}}            & \multicolumn{2}{c}{\textbf{Distinct Bigram}} \\ \cline{3-6} 
                                                                               &                                                                                & \multicolumn{1}{l|}{Supervised} & Diversity RL     & Supervised      & Diversity RL         \\ \hline
Beam Search with Trie                                                                      & -                                                                              & {0.590}                           & {0.678} & {0.868}           & {0.895}      \\ 
Diverse Sibling Search with Trie                                                        & 0.1                                                                            & {0.591}                           & {0.706} & {0.869}           & {0.903}      \\ 
Diverse Sibling Search with Trie                                                        & 0.5                                                                            & {0.616}                           & {0.716} & {0.876}           & {0.908}      \\ 
Diverse Sibling Search with Trie                                                        & 1                                                                              & {0.651}                           & \textbf{0.738} & {0.889}        & \textbf{0.917}     \\ \hline
\end{tabular}
\caption{Comparison of lexical diversity b/w supervised finetuning \& CLOVER (diversity driven RL) for various decoding algorithms}
\label{tab:result_diveristy_comparison}
\end{table*}
\subsection{Evaluation Metrics}
Our objective is to generate rewrites which (i) are valid bid keywords, (ii) are of high quality, (iii) have high coverage. As we discussed in section \ref{evaluation_model}, heuristic-based evaluation metrics such as BLEU \cite{bleu}, ROUGE \cite{rouge} that rely on n-gram overlap perform poorly in evaluating the rewrite quality. Hence, we use our trained evaluation model described in section \ref{evaluation_model} to evaluate the rewrite or keyword quality. Our evaluation model achieves AUC-ROC of 0.924 and 0.910 on the English and Non-English human judgments data (test split), respectively. Apart from the three criteria mentioned above, there is another important criterion specific to offline query rewriting. In the offline setting, the production table already consists of high-quality keywords retrieved by various algorithms proposed in the literature. Therefore, the usefulness of a newly proposed algorithm depends on how many net new high-quality bid keywords an algorithm adds to the production table. In other words, an algorithm creates no additional value in generating keywords that have already been retrieved and stored in the production table by previous methods. We apply a series of filters on the generated rewrites to obtain the net new high-quality bid keywords an algorithm adds to the production table:

\begin{itemize}
    \item \textbf{Valid Bid Filter:} We filter out all the generated rewrites that are not present in the bid keyword library. 
    
    \item \textbf{Unique Filter: } We filter out those keywords that are already present in the production map \textit{i.e.} we only select those keywords that are unique to the algorithm and not present in the production map. 
    
    \item \textbf{Quality Filter (Global):} Here, we filter out the generated keywords with a quality score less than a particular threshold. A common threshold is chosen across all the input queries; hence we call this a global filter. This filter ensures we only select keywords that have a global quality standard. 
    
    \item \textbf{Quality Filter (Local):} We additionally apply a filter using a quality threshold that is specific to each query. We compute the minimum quality of the keywords present in the production map for each query and use it as the (query-specific) quality threshold. Hence, we ensure that the newly added keywords have a higher quality than the minimum quality keyword present in the production dictionary for each query. 
\end{itemize}

Finally, we use the following evaluation metrics: 

\textbf{Unique High-Quality Bid Keyword} refers to the total number of keywords retrieved after applying all the filters mentioned above. We report the average value per input query.

\textbf{Distinct n-gram statistics} \cite{ diversity_mutual_information} measure the diversity in the rewrites after applying all the filters. It counts the total number of distinct n-grams in the generated rewrites. The statistics are then divided by the total number of tokens generated to bias against long rewrites.


\subsection{Results \& Discussions}
\label{sec:offline_results}

We report our offline evaluation results on various algorithms in table \ref{tab:OfflineResults}. The first row contains the results of using beam search without the prefix trie during inference. In all the other experiments, we used trie based decoding. We use a beam size of 20 in all these experiments. The reported results are aggregated across the different languages. The performance across the different languages are similar, and hence we do not include the language-wise data due to space constraints. As noted by \cite{baidu_generative_trie}, using a prefix trie
during beam search significantly improves the number of new keywords added to the production map because the trie constraint ensures that all the generated rewrites are valid bid keywords, hence serves as a strong baseline. However, we note that the lexical diversity of the rewrites, as measures by the distinct n-gram statistics, are low for the rewrites generated with beam search. In the next row, we report the result of using diverse sibling search \cite{DSS} instead of normal beam search during inference on the supervised finetuned model. We tuned the diversity strength ($\gamma$) hyper-parameter to maximize the number of unique, high-quality keywords retrieved. We observe slight improvements in (i) the number of keywords retrieved (from 3.80 to 3.84) and (ii) the diversity of the generated rewrites. However, as we show in the ensemble results (Ensemble (I) + (II) + (III)), out of the 3.84 keywords/query only an average of 0.39 keywords/query are the net new additions to the production map because the remaining keywords were already retrieved using standard beam search. 

\begin{figure}[h]
    \centering
    \includegraphics[scale=0.5,width=0.8\linewidth]{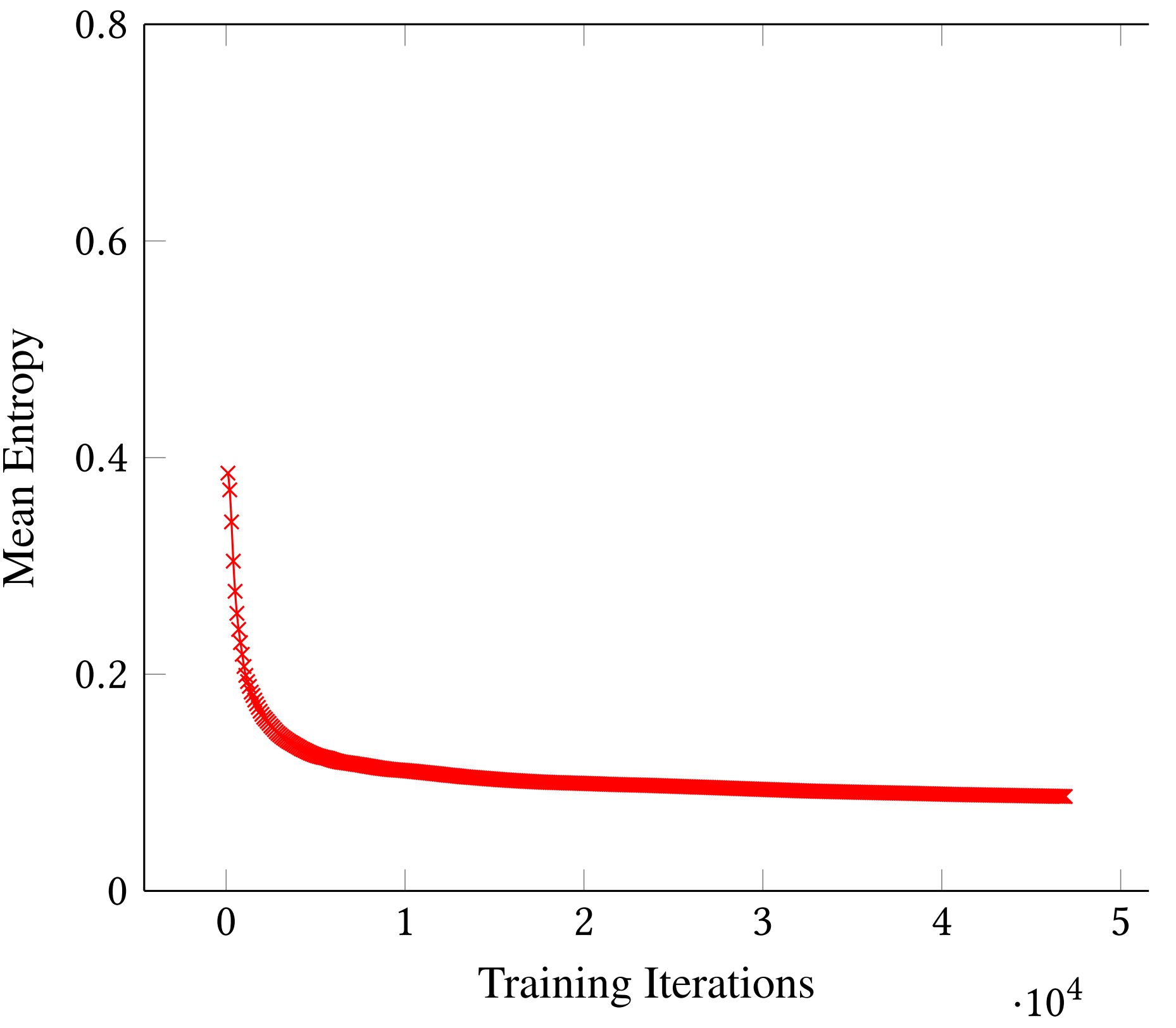}
    \caption{Entropy of the Output distribution as a function of the training iterations in the standard reinforcement learning setup (REINFORCE with a baseline algorithm)}
    \label{fig:entropy_graph}
\end{figure}

In figure \ref{fig:entropy_graph}, we show the entropy of the output distribution as a function of the training iterations during standard RL finetuning. We observe that the entropy does indeed decreases to very low values as discussed in section \ref{diversity_driven_rl}. As a result, we observed that the coverage of the rewrites significantly deteriorates since the output distributions are very peaky. We then report the results of the model finetuned using our proposed diversity driven RL approach in table \ref{tab:OfflineResults}. We observe a significant improvement in the diversity of the rewrites while the number of retrieved keywords is still comparable to those retrieved by previous methods. Further, as shown in the last row, most rewrites from the diversity RL (CLOVER) method (2.16 out of 3.13) are net new additions to the production map. Here, we wish to note that generating lexically diverse keywords is much more challenging than generating lexically similar keywords. For instance, consider the example in table \ref{tab:DiversityExample} where the input query is "accommodations in london". The lexically similar rewrites generated by supervised finetuning approaches such as "accommodations London", "london accommodation" are often just minor syntactic variations while the generations using CLOVER such as "best hotels london", "lodging in london uk" require a greater understanding of the search intent. In table \ref{tab:DiversityExample}, we compare the rewrite diversity between supervised and our proposed diversity RL approach for various decoding algorithms. We show that the rewrite diversity is significantly higher in CLOVER across the different decoding algorithms. Finally, in table \ref{tab:beam_size_comparison}, we increase the beam size during inference and show that both the number of retrieved keywords and their diversity increase in our proposed approach. 


\begin{table}[]
\begin{tabular}{cccc}
\hline
Beam Size & \begin{tabular}[c]{@{}c@{}}Unique High Quality\\  Bid Keywords\end{tabular} & \begin{tabular}[c]{@{}c@{}}Distinct \\ Unigram\end{tabular} & \begin{tabular}[c]{@{}l@{}}Distinct \\ Bigram\end{tabular} \\ \hline
20        & 3.13                                                                        & 0.738                                                       & 0.917                                                      \\
30        & 3.96                                                                        & 0.753                                                       & 0.924                                                      \\ 
40        & 4.62                                                                        & 0.771                                                       & 0.932                                                      \\ 
50        & \textbf{5.14}                                                                        & \textbf{0.785}                                                       & \textbf{0.937}                                                      \\ \hline
\end{tabular}
\caption{Effect of varying the Beam Size in our generator model finetuned with diverse RL objective and decoded with diverse sibling search $(\gamma=1)$}
\label{tab:beam_size_comparison}	
\end{table}

\section{Online Experiment}

To measure the efficacy of CLOVER, we conducted online experiments on live traffic of a large commercial search engine. The data was collected for more than two weeks over a significant percentage of traffic via the A/B testing framework.
We use two primary metrics to evaluate the performance of our retrieval framework. 

\begin{itemize}
    \item \textbf{Click} which denotes the net new click addition over the existing rewrite models. This translates into revenue.
    \item \textbf{Defect Quality} - This data is collected via human-labeled judgment where we sent the impressed query and keyword pairs each day for a month from both production and experiment setting to be categorized into good and bad pairs. The delta change represents the decrease in bad labeled data with respect to all data collected.
\end{itemize}

We chose the above two metrics to ensure that click/revenue increase should not come at the cost of quality deterioration. Table \ref{tab:OnlinResults} reflects the summarized results of the stated metric for both English and non-English queries. 










\begin{figure}
    \centering
    \includegraphics[width=1.0\linewidth]{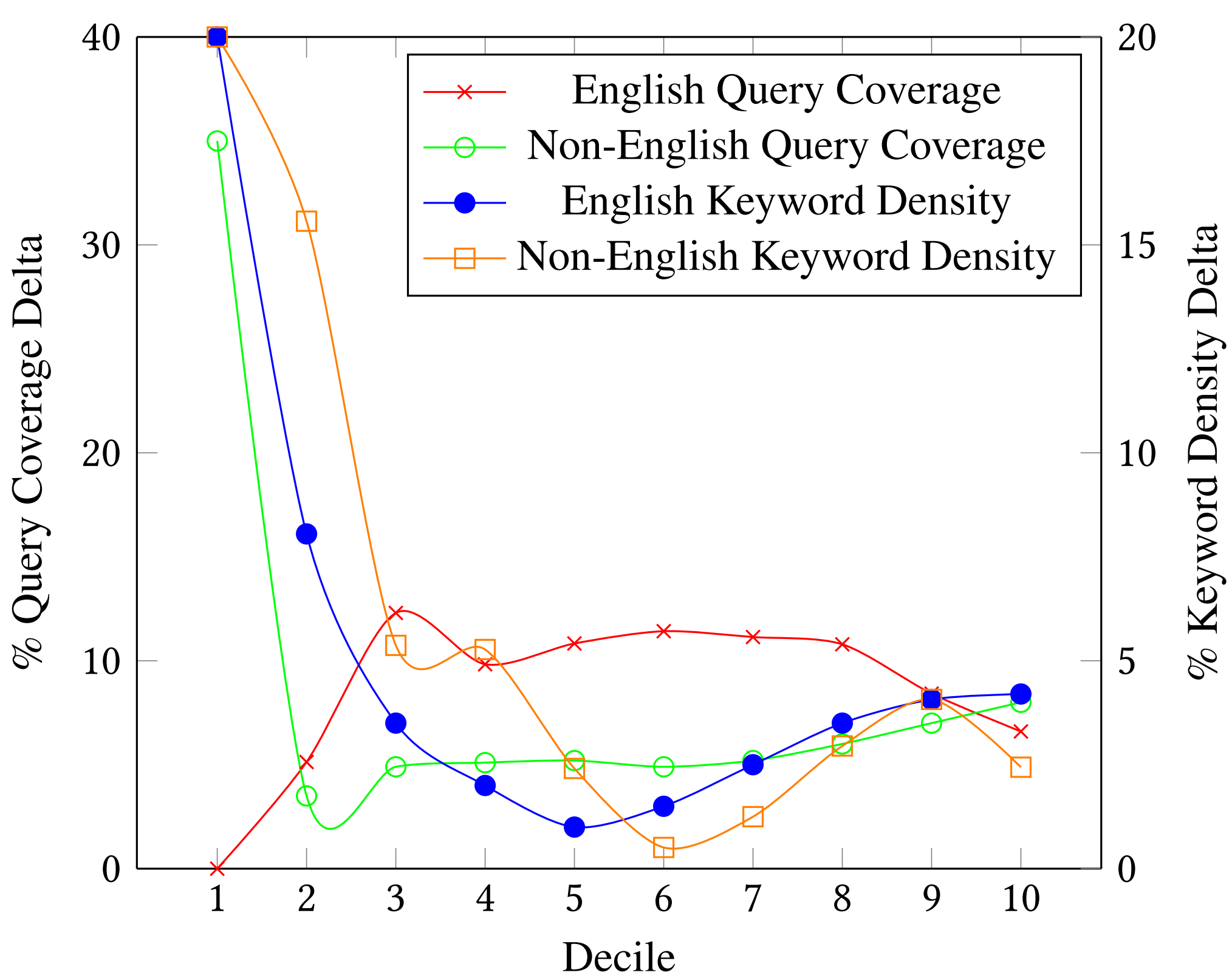}
    \caption{Coverage (left y-axis) and Density (right y-axis) delta uplift from current production map over each decile}
    \label{fig:online_graph}
\end{figure}

CLOVER contributed to a healthy increase of clicks (~8\% in non-English and ~18\% in English segments) without compromising on the quality of rewrites generated. In fact, we are able to improve the defect rate by a significant margin (~-18\% in non-English and ~-9\% in English segments). To further investigate the gains, we analyzed the query coverage and keyword density increase over the current production map. We distributed queries into buckets (otherwise commonly known as deciles in the field of sponsored search); each bucket consists of queries depending on the frequency in which they occur. For example:- decile 1 represents queries that are most frequent in number, but the overall count of such queries is less (usually under 100), similarly owing to the heavy tail nature of search queries, decile 10 represents queries that are less frequent, but overall distinct queries are very high in number (usually in millions). We define \textbf{query coverage} for a decile as the number of queries in the decile for which any sponsored content was shown, also \textbf{keyword density} for a query is defined as the number of keywords that is mapped to a particular query in the production map. Figure \ref{fig:online_graph} shows that for both English and non-English languages, we see a healthy query coverage increase leading to relevant sponsored content on searches which were earlier untapped. We also observe a healthy keyword density increase for existing queries leading to more relevant and better service of sponsored content, supported by improved defect metric.
\begin{center}
\begin{table}[h]	
	\begin{tabular}{c c c c}		
		\hline
		 Language Segment & $\Delta$ Click &$\Delta$ Revenue  & $\Delta$ Defect \\ \hline
        Non-English & 7.90\%  & 16.47\% & -18.47\% \\
        English & 17.76\%  &26.11\% & -9.48\% \\ \hline
	\end{tabular}	
	\caption{Online Results for A/B Testing on live traffic}
	\label{tab:OnlinResults}	
\end{table}
\end{center}
\section{Conclusion}
Generating high-quality, diverse rewrites of search queries is a critical challenge in sponsored search and can lead to significant revenue gains.
The existing generative Seq2Seq models do not yield satisfactory results due to disparity in training objectives and desired outcomes. In this paper, we analyze the challenges of existing models and propose a Close Variant Generation (CLOVER) framework to optimize human judgments on rewrite quality with diversity constraints on the rewrite.
We show that our proposed approach generates diverse and high-quality keywords that were not retrieved by previous methods. Through live A/B tests on a popular search engine, we show significant gains in clicks, revenue, defect, and other metrics over state-of-the-art techniques currently in production.


\bibliographystyle{ACM-Reference-Format}
\bibliography{references}


 



\end{document}